\documentclass{article} 
\usepackage[preprint]{colm2025_conference}

\usepackage{microtype}
\usepackage{hyperref}
\usepackage{url}
\usepackage{booktabs}
\usepackage{natbib}
\usepackage{lineno}
\usepackage{graphicx}
\usepackage{wrapfig}
\usepackage{tabularx}
\usepackage[edges]{forest}
\usepackage{xcolor}

\usepackage{tikz}
\usetikzlibrary{shapes.geometric, arrows.meta, positioning}

\tikzstyle{mainstage} = [rectangle, rounded corners, minimum width=1.2cm, minimum height=1cm, text centered, draw=black, fill=blue!20, rotate=90]
\tikzstyle{substage} = [rectangle, rounded corners, minimum width=3.5cm, minimum height=1cm, text centered, draw=black, fill=gray!10]
\tikzstyle{arrow} = [thick, ->, >=stealth]

\definecolor{darkblue}{rgb}{0, 0, 0.5}
\hypersetup{colorlinks=true, citecolor=darkblue, linkcolor=darkblue, urlcolor=darkblue}

\title{A Survey on Hypothesis Generation for Scientific Discovery in the Era of Large Language Models}

\author{Atilla Kaan Alkan$^{1,11}$, Shashwat Sourav$^{2,11}$, Maja Jablonska$^{3,11}$, Simone Astarita$^{4,11}$, \And Rishabh Chakrabarty$^{5,11}$, Nikhil Garuda$^{6,11}$, Pranav Khetarpal$^{7,11}$, Maciej Pióro$^{8,11}$, \And Dimitrios Tanoglidis$^{9,11}$, Kartheik G. Iyer$^{10,11}$, Mugdha S. Polimera$^{1,11}$, Michael J. Smith$^{11}$, \And Tirthankar Ghosal$^{12,11}$, Marc Huertas-Company$^{13,11}$, Sandor Kruk$^{14,11}$, \And Kevin Schawinski$^{15,11}$ \& Ioana Ciucă$^{16,11}$ \\ \\ 
$^{1}$ Center for Astrophysics, Harvard \& Smithsonian, Cambridge, MA, USA \\
$^{2}$ Washington University in St. Louis \\
$^{3}$ Australian National University \\
$^{4}$ European Commission, Joint Research Centre (JRC) \\
$^{5}$ Intelligent Internet Inc. \\
$^{6}$ University of Arizona \\
$^{7}$ Indian Institute of Technology, Delhi \\
$^{8}$ Institute of Fundamental Technological Research, Polish Academy of Sciences \\
$^{9}$ Walgreens Boots Alliance AI Lab \\
$^{10}$ Columbia University \\
$^{11}$ UniverseTBD \\
$^{12}$ Oak Ridge National Laboratory \\
$^{13}$ Instituto de Astrofísica de Canarias \\
$^{14}$ European Space Agency \\
$^{15}$ Modulos AG \\
$^{16}$ Stanford University \\
}

\tikzset{
    block/.style = {draw, rectangle, rounded corners, fill=blue!10, text width=3cm, align=center, minimum height=1cm},
    method/.style = {draw, rectangle, rounded corners, fill=purple!10, text width=2.8cm, align=center, minimum height=1cm},
    arrow/.style = {->, thick, >=Stealth}
}

\begin{document}

\ifcolmpreprint
\fi

\maketitle

\begin{abstract}
Hypothesis generation is a fundamental step in scientific discovery, yet it is increasingly challenged by information overload and disciplinary fragmentation. Recent advances in Large Language Models (LLMs) have sparked growing interest in their potential to enhance and automate this process. This paper presents a comprehensive survey of hypothesis generation with LLMs by (i) reviewing existing methods, from simple prompting techniques to more complex frameworks, and proposing a taxonomy that categorizes these approaches; (ii) analyzing techniques for improving hypothesis quality, such as novelty boosting and structured reasoning; (iii) providing an overview of evaluation strategies; and (iv) discussing key challenges and future directions, including multimodal integration and human-AI collaboration. Our survey aims to serve as a reference for researchers exploring LLMs for hypothesis generation.
\end{abstract}

\section{Introduction}
Hypothesis generation is a fundamental component of scientific discovery, enabling researchers to formulate testable predictions and uncover new insights. This process has traditionally relied on human intuition, experience, and domain expertise. However, as the volume of scientific literature grows drastically, researchers face challenges in assimilating relevant knowledge across disciplines. This information saturation creates bottlenecks that hinder the discovery of new insights.

From a philosophy of science perspective, a hypothesis can be defined as a tentative explanation or prediction about a phenomenon, formulated in a way that allows for empirical testing and potential falsification~\citep{popp-logi}. Despite its crucial role in the scientific method, hypothesis generation remains constrained by disciplinary silos and cognitive overload. Traditional approaches struggle to integrate knowledge across fields, limiting researchers’ ability to identify interdisciplinary connections that may lead to groundbreaking discoveries.


In this context, generative Large Language Models (LLMs) such as GPT~\citep{Radford2018ImprovingLU}, PaLM~\citep{Chowdhery2022PaLMSL}, LLaMA~\citep{Touvron2023LLaMAOA}, and Mistral~\citep{Jiang2023Mistral7} have emerged as promising systems to overcome these barriers. By leveraging vast repositories of scientific texts, LLMs can process, synthesize, and generate novel hypotheses, assisting human expertise and facilitating interdisciplinary research. Since the introduction of LLMs, there has been a growing research interest in hypothesis generation using these models, as illustrated in. The number of research papers on this topic has significantly risen, highlighting the increasing recognition of LLMs' potential in scientific exploration.


Despite their promise, LLM-driven hypothesis generation presents several challenges. Evaluating generated hypotheses remains a complex issue, requiring novelty, relevance, feasibility, significance, and clarity assessment. A major concern is ensuring that LLMs generate innovative hypotheses rather than paraphrasing existing knowledge. Furthermore, the quality and diversity of training data play a crucial role in the effectiveness of these models. Biases present in the datasets can influence the generated hypotheses, potentially reinforcing existing perspectives while overlooking unconventional or groundbreaking ideas. Furthermore, integrating LLMs into the scientific process requires addressing issues related to interpretability, reliability, and validation of machine-generated hypotheses.

In this paper, we aim to provide a comprehensive overview of the state of hypothesis generation using LLMs. We examine the current methodologies, categorize existing approaches into a structured taxonomy, and discuss the challenges and limitations inherent in this emerging field. By addressing these aspects, we seek to outline key research directions for the community on the generation of LLM-based hypotheses.

\section{Paper Collection Methodology}

We employed a systematic literature retrieval strategy combining keyword-based search and manual curation to construct a comprehensive and historically grounded survey of computational approaches to scientific hypothesis generation. The primary objective was to capture studies spanning both pre-LLM methods (e.g., literature-based discovery (LBD) and early NLP techniques) and more recent approaches involving LLMs.

\subsection{Search Strategy}

We queried the arXiv API to retrieve relevant publications for this survey paper. No explicit time range was imposed on the queries; the retrieved papers span a publication range from 2005 to 2025. To ensure domain relevance, we restricted the search to papers categorized under Computer Science, specifically within the \texttt{cs.CL} (Computation and Language) category. We used a curated list of search terms designed to reflect the thematic scope of our study, covering core concepts, traditional approaches, and recent methods involving natural language processing and large language models. The full list of search terms is presented in Table~\ref{tab:search-terms}.

\begin{table}[ht]

\centering

\begin{tabularx}{\linewidth}{>{\raggedright\arraybackslash}X >{\raggedright\arraybackslash}X >{\raggedright\arraybackslash}X}
\toprule
\textbf{Core Concepts} & \textbf{Recent Techniques} & \textbf{Traditional Techniques} \\
\midrule
\texttt{hypothesis generation} & \texttt{NLP for hypothesis generation} & \texttt{Swanson hypothesis discovery} \\
\texttt{scientific hypothesis generation} & \texttt{language models scientific discovery} & \texttt{open discovery system} \\
\texttt{scientific discovery} & \texttt{large language models hypothesis generation} & \texttt{ABC model literature discovery} \\
\texttt{automated scientific discovery} & \texttt{knowledge graph hypothesis generation} & \texttt{semantic predications scientific discovery} \\
& \texttt{question generation scientific research} & \texttt{semantic indexing hypothesis generation} \\
& \texttt{natural language processing scientific discovery} &\texttt{literature based discovery} \\
& \texttt{machine learning hypothesis generation} & \\
& \texttt{automated reasoning for discovery} & \\
& \texttt{discovery using LLMs} & \\
\bottomrule
\end{tabularx}
\label{tab:search-terms}
\caption{Search terms used for systematic retrieval, grouped by theme.}
\end{table}

\subsection{Inclusion Criteria}
Since our search was conducted exclusively via the arXiv API, the articles retrieved are primarily pre-prints. However, many of these works may have subsequently appeared in peer-reviewed journals, conference proceedings, or workshop venues. We included all papers that satisfied at least one of the following criteria, regardless of their publication status at the time of retrieval:

\begin{itemize}
    \item The paper proposes or evaluates an automated or semi-automated method for scientific hypothesis generation;
    \item The work addresses scientific discovery through natural language processing, knowledge graph mining, or large language models;
    \item The paper contributes theoretical insights or historical perspectives on scientific discovery, particularly regarding the role of AI in hypothesis formulation.
\end{itemize}

\subsection{Review Process}
After initial filtering based on titles and abstracts, the remaining set of papers was manually screened for relevance and categorized according to methodological paradigm (e.g., LBD, NLP, LLMs, hybrid systems), scientific domain (e.g., biomedicine, astrophysics, chemistry), and hypothesis representation. This classification enabled us to trace the evolution of techniques and the shifts in hypothesis formalization over time.

\section{Methods for Scientific Hypothesis Generation}
\begin{figure}[ht]
\centering
\begin{tikzpicture}[node distance=1cm and 1.5cm]

    \node (human) [substage, fill=blue!20] {Human-Centric};
    \node (lbd) [substage, fill=red!20] {LBD};
    \node (textmining) [substage, fill=orange!20, below=of lbd] {Text-Mining};
    \node (supervised) [substage, fill=green!20, below=of textmining] {Supervised Learning};
    \node (graph) [substage, fill=cyan!20, below=of supervised] {Graph Based};
    \node (llms) [substage, fill=purple!10, below=of graph] {LLM-Driven};

        \node (arrowprompting) [substage, fill=purple!10, right=1.8cm of llms, yshift=1.8cm] {Prompting};
        \node (arrowfinetuning) [substage, fill=purple!10, below=of arrowprompting] {Fine-Tuning};
        \node (arrowrag) [substage, fill=purple!10, below=of arrowfinetuning] {RAG};
        \node (arrowkg) [substage, fill=purple!10, below=of arrowrag] {Knowledge Graphs};
        
        \draw [arrow] (llms.east) -- ++(0.5,0) |- (arrowprompting.west);
        \draw [arrow] (llms.east) -- ++(0.5,0) |- (arrowfinetuning.west);
        \draw [arrow] (llms.east) -- ++(0.5,0) |- (arrowrag.west);
        \draw [arrow] (llms.east) -- ++(0.5,0) |- (arrowkg.west);

    \path (lbd.north) -- (llms.south) coordinate[midway] (centerpoint);

    \node (main) [mainstage, left=3.5cm of centerpoint] {SHG};

     \draw [arrow] (main.south) -- ++(0.5,0) |- (human.west);
    \draw [arrow] (main.south) -- ++(0.5,0) |- (lbd.west);
    \draw [arrow] (main.south) -- ++(0.5,0) |- (textmining.west);
    \draw [arrow] (main.south) -- ++(0.5,0) |- (supervised.west);
    \draw [arrow] (main.south) -- ++(0.5,0) |- (graph.west);
    \draw [arrow] (main.south) -- ++(0.5,0) |- (llms.west);

\end{tikzpicture}
\caption{Taxonomy of Methods for Scientific Hypothesis Generation (SHG).}
\label{fig:taxonomy}
\end{figure}
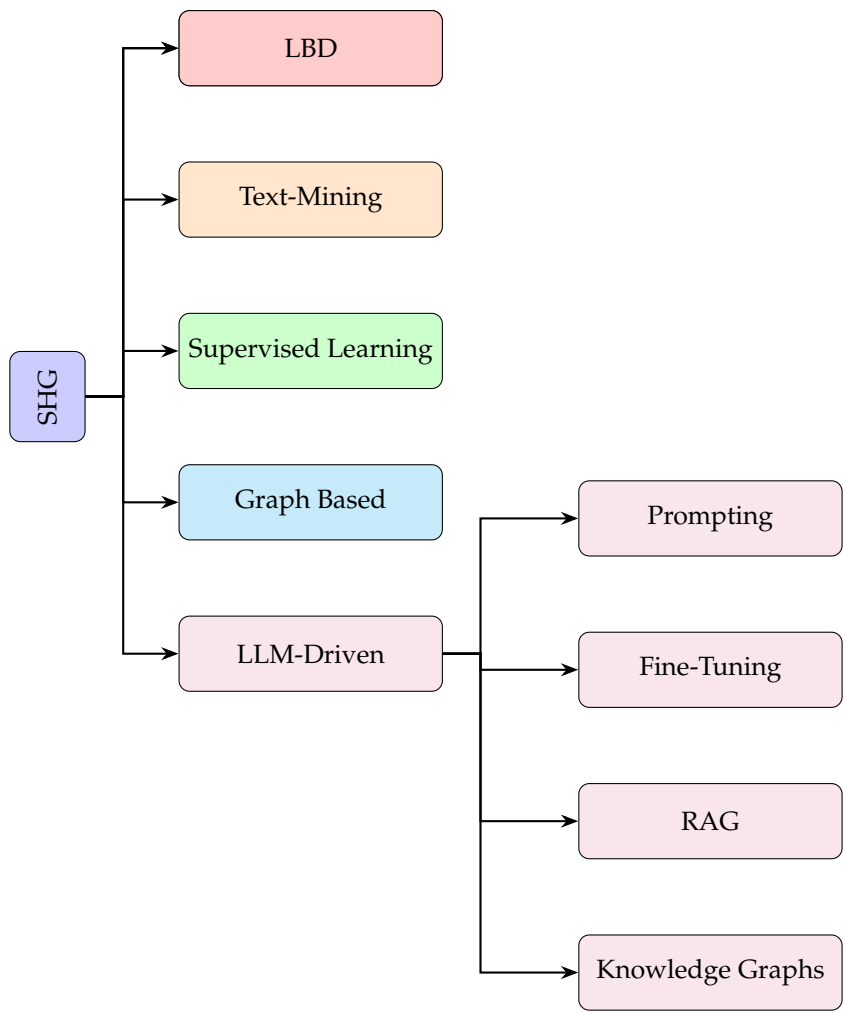

In this section, we review the existing methods for generating scientific hypotheses. As illustrated in Figure~\ref{fig:taxonomy}, we present methods ranging from early approaches such as literature-based discovery (LBD), text mining, and statistical learning methods to more recent techniques, including graph-based models and LLM.


\subsection{Human-Centric}
\label{humancentric}

Human-centric methods are based on researchers' expertise, intuition, and theoretical and practical knowledge. In this approach, individual insights gained through years of research and practical exposure play an important role in forming new hypotheses. Researchers engage in brainstorming sessions and discussions, where their knowledge and observational skills help identify new trends and anomalies \citep{swanson1986fish,nonaka2009knowledge}. This method takes into account the ability of an individual to grasp context-specific nuances and reinterpret existing information. However, this method can have cognitive biases and the risk of overreliance on conventional paradigms, which may limit the exploration of less familiar or interdisciplinary ideas.

\subsection{Literature-based Discovery}
\label{lbd}
Literature-based discovery (LBD) leverages computational tools to mine the vast scientific literature for implicit or previously overlooked connections between concepts not directly linked in published research. The foundational idea of LBD introduced by~\citet{swanson1986undiscovered}, relies on the notion of "undiscovered public knowledge"—information that exists in the literature but remains unconnected due to disciplinary silos or the overwhelming volume of publications. The author's seminal study, which identified a potential link between fish oil and Raynaud’s syndrome by connecting disparate literature, remains a landmark example of this approach. Over time, LBD has evolved to incorporate advances in text mining, natural language processing (NLP), and semantic analysis, enabling the automated discovery of hidden relationships at scale~\citep{smalheiser2017rediscovering}. Several systems have been developed to operationalize LBD. One of the earliest tools, \texttt{ARROWSMITH}~\citep{SMALHEISER1998149}, identifies intermediate terms (B) that link two disjoint sets of articles (A and C), thereby suggesting novel hypotheses. More recent systems have expanded the methodological toolkit. \texttt{MOLIERE}~\citep{sybrandt2017moliereautomaticbiomedicalhypothesis}, for example, builds semantic networks from MEDLINE and other biomedical data using topic modelling techniques such as Latent Dirichlet Allocation~\citep{10.5555/944919.944937_lda} and phrase mining to uncover latent themes and suggest short conceptual paths between topics. \texttt{KnIT}~\citep{knit} is another system that extracts factual statements from the literature, represents them in a queryable network and applies information diffusion algorithms to generate hypotheses, such as discovering novel kinases that phosphorylate p53. Other approaches rely on structured vocabularies like the Medical Subject Headings (MeSH, \citep{lipscomb2000msh}) to build profile-based representations of concepts and identify indirect links between them. Systems like \texttt{DiseaseConnect}~\citep{diseaseconnect} and \texttt{BrainSCANr}~\citep{VOYTEK201292} also apply text mining to biomedical abstracts to uncover latent semantic features and generate novel associations. LBD remains especially valuable in information-rich domains, where human researchers may miss non-obvious connections across disciplines. By systematically surfacing these links, LBD tools support cross-domain discovery and can help accelerate hypothesis formulation in complex scientific landscapes.

\paragraph{Supervised Learning}
\label{statiscal and discovery}

Statistical computing methods are robust and data-driven for the hypothesis generation task. This set of methods uses statistical tools such as regression, clustering and Bayesian inference. On the other hand, computational approaches, including machine learning algorithms and network analysis, extend these tools by handling higher dimensional data and investigating complex, multivariate relations that may exist in the data. Some of the recent works \citep{breiman2003statistical}.

\subsection{LLM-Driven}
\label{hypogen_with_llms}

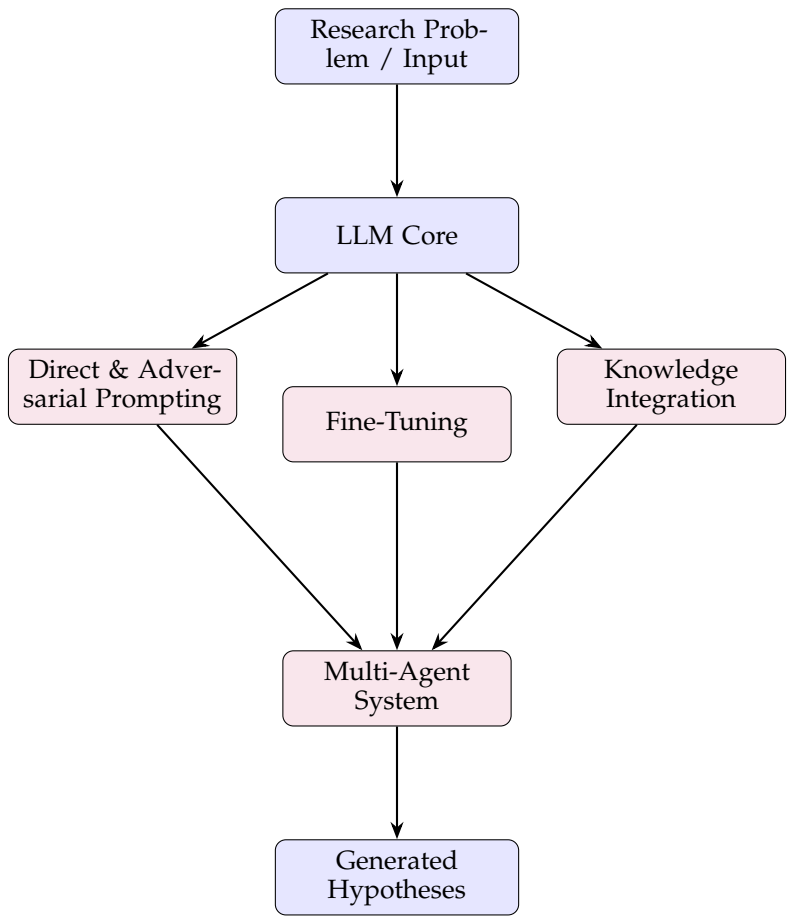
\begin{figure}[!h]
\centering
\begin{tikzpicture}[node distance=1.5cm]
    \node (input) [block] {Research Problem / Input};
    \node (llmcore) [block, below=of input] {LLM Core};
    
    \node (prompting) [method, below left=1cm and 0.5cm of llmcore] {Direct \& Adversarial Prompting};
    \node (finetune) [method, below=of llmcore] {Fine-Tuning};
    \node (knowledge) [method, below right=1cm and 0.5cm of llmcore] {Knowledge Integration};
    
    \node (multiagent) [method, below=of finetune, yshift=-1cm] {Multi-Agent System};
    
    \node (output) [block, below=of multiagent] {Generated Hypotheses};
    
    \draw [arrow] (input) -- (llmcore);
    \draw [arrow] (llmcore) -- (prompting);
    \draw [arrow] (llmcore) -- (finetune);
    \draw [arrow] (llmcore) -- (knowledge);
    \draw [arrow] (prompting) -- (multiagent);
    \draw [arrow] (finetune) -- (multiagent);
    \draw [arrow] (knowledge) -- (multiagent);
    \draw [arrow] (multiagent) -- (output);
\end{tikzpicture}
\caption{Pipeline of LLM-Driven Hypothesis Generation. The process begins with a research problem, which is processed by the LLM core. Various methodological branches (e.g., Direct \& Adversarial Prompting, Fine-Tuning, and Knowledge Integration) contribute to a multi-agent framework that converges to generate hypotheses.}
\label{fig:llm_pipeline}
\end{figure}

Large Language Models (LLMs) have recently emerged as powerful tools for scientific hypothesis generation. Figure~\ref{fig:llm_pipeline} summarizes the pipeline and the common techniques used to generate hypothesis Their capacity to process vast corpora of scientific texts and synthesize information makes them particularly well-suited to this task. The literature reveals many methods that leverage LLMs to assist in or fully automate the generation of scientific hypotheses. These methods can be categorized as follows:

\paragraph{Direct and Adversarial Prompting}  
Direct prompting involves formulating clear and concise instructions to perform hypothesis generation directly using an LLM. In this method, users design a prompt explicitly asking the model to propose potential explanations or predictions based on a given context. One can ask the LLM a question or give it direct instruction about a topic, and it responds with possible ideas or explanations. This approach benefits from its simplicity and ease of implementation, allowing researchers to quickly gauge the model's capacity for innovative reasoning~\citep{Radford2018ImprovingLU}. However, the output quality highly depends on the prompt's clarity and the model's inherent understanding of the subject matter. The adversarial prompting approach is designed to make the LLM go beyond its standard response patterns by introducing counterfactual or challenging scenarios. By deliberately framing prompts in a way that exposes the model to unconventional perspectives, researchers can encourage the generation of hypotheses that diverge from common assumptions. This method can involve contrasting ideas or setting up dilemmas that force the model to explore unexplored paths of reasoning. Adversarial prompting not only tests the robustness of the LLM but also helps in detecting biases inherent in its training data, ultimately leading to more diversified and potentially groundbreaking insights~\citep{chowdhery2023palm}.

\paragraph{Fine-tuning}  
Rather than relying solely on prompting, this method involves fine-tuning an LLM on domain-specific datasets containing both foundational knowledge and corresponding hypotheses extracted from the literature. This tuning process enables the model to learn the patterns and context typically associated with hypothesis formulation. In one study, a temporally split biomedical dataset was used to test the model’s capacity to generate plausible hypotheses after fine-tuning.


\paragraph{Knowledge Integration}  
Some methods incorporate structured knowledge from scientific knowledge graphs to improve relevance and reduce hallucinations. These graphs encode entities and their relationships, which serve as grounding information for the LLM. A representative approach, KG-CoI (Knowledge-Grounded Chain of Ideas), uses graphs for context retrieval, chain-of-thought generation, and hallucination detection, improving the reliability of generated hypotheses \citep{kgcoi}. Integrating knowledge in this form can help overcome some shortcomings of classical text-based RAG (Retrieval-Augumented Generation), such as the potential omission of rare but crucial information. Graph structures can also capture causal relationships between concepts and prove useful when generating hypotheses bridging selected concepts, as demonstrated by \cite{causal_graphs}. 


\paragraph{Multi-Agent System}  
This approach introduces multiple LLM agents with different roles -such as analyst, scientist, or critic - that interact to collaboratively generate and evaluate hypotheses. Through dialogue and feedback between agents, this framework aims to produce more innovative and better-grounded scientific ideas.

\newpage
\section{Evaluation Methodologies}
Evaluating systems for scientific hypothesis generation is a complex task. Unlike traditional NLP evaluation, hypothesis generation aims to produce novel, plausible, and testable scientific ideas—often in domains where ground truth is incomplete or non-existent. This open-endedness renders standard evaluation metrics insufficient and necessitates a multi-faceted approach combining human expertise, automated metrics, multi-modal integration, and domain-specific validation. In this section, we first review established methodologies before outlining promising directions for future research.

\subsection{Human Expert Evaluation}

\paragraph{Expert Assessment Protocols}
Evaluations conducted by domain experts remain the most reliable method for assessing the relevance, originality, and scientific merit of machine-generated hypotheses. Over time, these assessments have become more structured and methodologically rigorous. Recent protocols have involved large panels of experts from diverse academic backgrounds to evaluate hypotheses along dimensions such as clarity, innovation potential, and expected impact. Comparative studies have shown that, when supported by LLMs, researchers can generate more compelling and diverse ideas than with traditional search-based workflows. Such findings suggest that expert-in-the-loop systems not only support hypothesis refinement but can also enhance ideation itself.

In highly specialized fields such as biomedicine, structured evaluations have been designed to focus on clinical relevance and biological plausibility. Frameworks developed for this purpose often involve expert reviews centred on real-world applicability and potential translational impact. Some benchmark efforts have incorporated expert assessments across multiple research tasks, offering a broader view of how LLMs contribute to domain-specific scientific workflows.

\paragraph{Blind Review and Pairwise Comparison}
To reduce bias and ensure fair evaluation, blind review protocols are increasingly employed. In these settings, experts are unaware whether a human or an AI system has generated a hypothesis. This approach has revealed that, in many cases, AI-generated hypotheses can be as highly rated—or even surpass—those written by human researchers regarding novelty and scientific interest. Building on this principle, some recent evaluation strategies employ direct pairwise comparisons in tournament-style formats, where hypotheses compete against each other and are ranked based on expert preference. These structured comparison schemes offer a scalable and interpretable method for evaluating generative systems.

\paragraph{Multi-Rater Reliability}
One of the persistent challenges in expert-based evaluation is achieving consistency across annotators. Scientific hypothesis assessment often involves subjective judgment, leading to variability in ratings. Earlier studies have highlighted relatively low agreement levels among reviewers, emphasizing the complexity of the task. However, newer frameworks are addressing this by introducing more formalized scoring rubrics, multiple rounds of review, and collaborative assessment protocols. These improvements have contributed to more stable and reproducible evaluation outcomes, reflecting a growing understanding of effectively integrating human judgment into validating AI-generated scientific content.

\subsection{Automated Evaluation}

\paragraph{Text-based Relevance}
Initial efforts to evaluate LLM outputs relied heavily on surface-level metrics such as BLEU and ROUGE, which measure word overlap between generated and reference hypotheses. However, such metrics often fall short of capturing the semantic depth and scientific value of an idea. As a result, more sophisticated evaluation tools have been developed that incorporate semantic precision and recall, as well as hybrid scores that combine symbolic and neural representations. These allow for a more meaningful assessment of whether a hypothesis is contextually appropriate and scientifically relevant. Additionally, some benchmarks now include domain-specific metrics tailored to the complexity and requirements of particular research tasks, such as code execution or model reproducibility.

\paragraph{Model-Based Metrics}
Recent evaluation frameworks have increasingly turned to large language models as evaluators of generated hypotheses. When fine-tuned or provided with structured prompts, these models can approximate human-level assessments across dimensions such as plausibility, novelty, and relevance. Some systems now rely on LLMs to score hypotheses using composite metrics that account not only for internal coherence but also for broader scientific context. For instance, measures have been developed to quantify how dissimilar a proposed idea is from past knowledge and how closely it aligns with emerging literature trends, thus reflecting historical uniqueness and prospective impact.

\paragraph{Novelty Assessment}
Measuring novelty remains one of the central goals in hypothesis evaluation. Automated approaches have evolved to estimate the originality of ideas by analyzing their semantic distance from existing publications. This often involves embedding-based comparisons using pre-trained scientific language models combined with ranking strategies that assess the rarity or innovation of proposed connections. Some systems build structured citation graphs or ideation chains to contextualize a hypothesis within a broader intellectual lineage, enabling more informed judgments about its uniqueness.

\paragraph{Domain-Specific Evaluation}
Evaluation strategies tailored to specific scientific fields are increasingly recognized as essential due to the varied standards of evidence, feasibility, and validation across disciplines. In biomedical research, hypothesis evaluation often relies on alignment with curated clinical databases or known gene-disease associations, enabling automated cross-referencing against structured biomedical knowledge. In the chemical sciences, evaluation protocols typically focus on structural validity and chemical plausibility, incorporating techniques such as molecular simulation or synthesis pathway prediction. Astronomy and astrophysics present unique challenges, where hypothesis evaluation may involve the integration of large-scale observational datasets or comparing generated hypotheses with complex knowledge graphs. Social science domains, on the other hand, prioritize theoretical grounding and temporal context, often requiring evaluation of whether a hypothesis is consistent with existing paradigms or predictive of future trends. These domain-specific practices underscore the importance of aligning evaluation methodologies with disciplinary norms, highlighting the need for adaptable frameworks that can accommodate the epistemological diversity of modern science.

\section{Challenges and Future Research Directions}
Despite substantial progress in developing LLM-based systems for scientific hypothesis generation, several critical challenges remain unresolved. One of the most pressing concerns is the issue of factual accuracy. LLMs are known to produce outputs that, while syntactically coherent and contextually plausible, can include erroneous or fabricated claims. This phenomenon, often referred to as hallucination, poses significant risks in scientific settings. Closely related is the challenge of interpretability. Most LLMs function as black-box systems, making it difficult to understand or trace the rationale behind specific hypotheses. This lack of transparency undermines trust and complicates the validation process, especially when hypotheses are intended to serve as the foundation for empirical research.

Bias is another persistent issue. Given that LLMs are trained on large, heterogeneous corpora, they tend to reproduce—and occasionally amplify—preexisting societal biases. These biases can influence the direction and framing of generated hypotheses, potentially skewing research priorities and excluding underrepresented perspectives. At the same time, the computational cost of training and deploying these models remains prohibitive for many institutions. The high energy and hardware requirements not only limit accessibility but also raise concerns about environmental sustainability. Domain adaptation poses additional hurdles. While fine-tuning on specialized datasets can enhance performance in specific fields, it often introduces the risk of overfitting, compromising the model’s ability to generalize across topics. Furthermore, the ethical implications of AI-generated hypotheses—from questions of authorship and accountability to the potential misuse of misleading hypotheses—remain largely unaddressed, necessitating the development of robust governance mechanisms.

To overcome these limitations, new methodological directions are emerging. Retrieval-augmented generation, which integrates LLMs with external scientific databases, offers a promising approach to grounding outputs in verifiable knowledge and reducing hallucinations. Another direction involves incorporating chain-of-thought reasoning or rationale tracing mechanisms, enabling models to generate not only hypotheses but also the reasoning pathways that led to their formulation. This increased transparency can help researchers evaluate the internal coherence and plausibility of generated ideas. Multi-agent collaborative frameworks are also gaining traction. Inspired by the collaborative nature of scientific inquiry, these systems simulate peer review or debate among virtual agents to refine and evaluate hypotheses dynamically. In the realm of fine-tuning, meta-learning and cross-domain transfer techniques are being explored to better balance specialization and generalization, allowing models to adapt flexibly to a variety of scientific domains without sacrificing rigour.

From a computational perspective, advances in model compression and energy-efficient architectures are expected to democratize access to LLM-based tools, making them more practical for research institutions with limited resources. At the same time, methodological co-design with ethicists, legal scholars, and domain experts is increasingly recognized as essential to developing socially responsible AI tools. Future systems should embed ethical safeguards, including bias detection, provenance tracking, and clear attribution protocols, directly into their design.

Complementing these methodological innovations, the field must adopt more sophisticated evaluation frameworks. Traditional metrics such as BLEU and ROUGE fall short of capturing hypotheses' semantic depth and scientific merit. In response, several novel evaluation paradigms are being developed. Scientific verifiability benchmarks, for instance, aim to assess whether generated hypotheses can be empirically tested or verified in real-world research. Temporal evaluation methods propose tracking the evolution of ideas over time—through citations, modifications, or integrations into published work—to assess long-term scientific impact. Evaluating the diversity and redundancy of generated hypotheses has also become a key area of interest, as the capacity to propose a broad range of novel ideas is a fundamental indicator of exploratory potential.

Another promising direction involves multi-modal evaluation, where hypotheses are assessed not just through text-based metrics but also through visualizations, structured knowledge graphs, or experimental data. Human-in-the-loop evaluation protocols are likewise gaining prominence. These frameworks involve iterative collaboration between researchers and models, enabling dynamic refinement and contextual validation of hypotheses. Finally, developing interdisciplinary evaluation standards is increasingly necessary as hypothesis-generation systems are deployed across various scientific domains. These standards must be flexible enough to accommodate domain-specific norms while preserving core principles such as novelty, relevance, verifiability, and scientific integrity.

In summary, while LLM-based systems have demonstrated considerable potential in augmenting scientific discovery, their limitations call for caution and innovation. Addressing persistent challenges such as factual inaccuracy, opacity, and domain sensitivity will require a coordinated effort across AI, domain science, and ethics. At the same time, the emergence of new methodological and evaluation paradigms offers a promising path toward developing robust, transparent, and impactful hypothesis-generation tools that align with the evolving standards of scientific research.

\section{Conclusion}

While Large Language Models have revolutionized the domain of automated text generation, their application in scientific hypothesis generation is still in its nascent stages and filled with challenges. Issues such as factual inaccuracies, lack of interpretability, inherent biases, and high computational demands underscore the need for continued research and innovation. This paper has reviewed the state-of-the-art methods for LLM-driven hypothesis generation and critically examined the accompanying limitations. Future research must prioritize the development of more transparent, efficient, and ethically sound models that can reliably support scientific inquiry. By addressing these challenges through interdisciplinary collaboration and methodological advances, the scientific community can unlock the full potential of LLMs, ultimately paving the way for transformative breakthroughs in knowledge discovery.

\bibliography{colm2025_conference}
\bibliographystyle{colm2025_conference}

\end{document}